%% file: main.tex
\documentclass[10pt,twocolumn,letterpaper]{article}

\usepackage[pagenumbers]{cvpr} 
\usepackage{multirow}
\usepackage{tabularx}
\definecolor{cvprblue}{rgb}{0.21,0.49,0.74}
\usepackage[pagebackref,breaklinks,colorlinks,allcolors=cvprblue]{hyperref}

\def\paperID{} 
\def\confName{}
\def\confYear{}

\title{PrivDFS: Private Inference via Distributed Feature Sharing against Data Reconstruction Attacks}

\author{
    Zihan Liu\textsuperscript{\rm 1}, Jiayi Wen\textsuperscript{\rm 1}, Junru wu\textsuperscript{\rm 1}, Xuyang Zou\textsuperscript{\rm 1}, Shouhong Tan\textsuperscript{\rm 1}, Zhirun Zheng\textsuperscript{\rm 2}, Cheng Huang\textsuperscript{\rm 1}\\
    \textsuperscript{\rm 1}College of Computer Science and Artificial Intelligence, Fudan University\\
    \textsuperscript{\rm 2}Department of Artificial Intelligence, Ajou University
}

\begin{document}
\maketitle
\input{0_abstract}    
\input{1_intro}

\input{2_background}

\input{3_method}
\input{4_evaluation}
\input{5_conclusion}

{
    \small
    \bibliographystyle{ieeenat_fullname}
    \bibliography{main}
}

\end{document}

%% file: 0_abstract.tex
\begin{abstract}
In this paper, we introduce PrivDFS, a distributed feature-sharing framework for input-private inference in image classification. A single holistic intermediate representation in split inference gives diffusion-based Data Reconstruction Attacks (DRAs) sufficient signal to reconstruct the input with high fidelity. PrivDFS restructures this vulnerability by fragmenting the representation and processing the fragments independently across a majority-honest set of servers. As a result, each branch observes only an incomplete and reconstruction-insufficient view of the input. To realize this, PrivDFS employs learnable binary masks that partition the intermediate representation into sparse and largely non-overlapping feature shares, each processed by a separate server, while a lightweight fusion module aggregates their predictions on the client. This design preserves full task accuracy when all branches are combined, yet sharply limits the reconstructive power available to any individual server. PrivDFS applies seamlessly to both ResNet-based CNNs and Vision Transformers. Across CIFAR-10/100, CelebA, and ImageNet-1K, PrivDFS induces a pronounced collapse in DRA performance, e.g., on CIFAR-10, PSNR drops from 23.25$\to$12.72 and SSIM from 0.963$\to$0.260, while maintaining accuracy within 1\% of non-private split inference. These results establish structural feature partitioning as a practical and architecture-agnostic approach to reducing reconstructive leakage in cloud-based vision inference.

\end{abstract}

%% file: 1_intro.tex
\section{Introduction}
\label{sec:intro}

Cloud-based inference, enabled by the rise of Machine Learning as a Service (MLaaS)~\cite{DBLP:conf/icmla/RibeiroGC15}, allows resource-constrained clients to access powerful AI models hosted in the cloud server. However, the deployment models expose raw client data---such as facial images, medical scans, and other sensitive inputs---to potentially untrusted servers, creating severe privacy risks~\cite{DBLP:journals/tai/RaoZWZSC25,DBLP:journals/tifs/OtroshiShahrezaHM24}. These risks highlight the need for private inference (PI), which aims to protect client inputs while maintaining the utility of the inference service~\cite{DBLP:journals/csur/MannWCB24}. In this work, we focus on PI for image classification, a dominant use case in cloud-based vision services where reconstruction attacks are particularly well-defined and strongly impactful.

\begin{figure}[t]
    \centering
    \includegraphics[width=0.9\columnwidth]{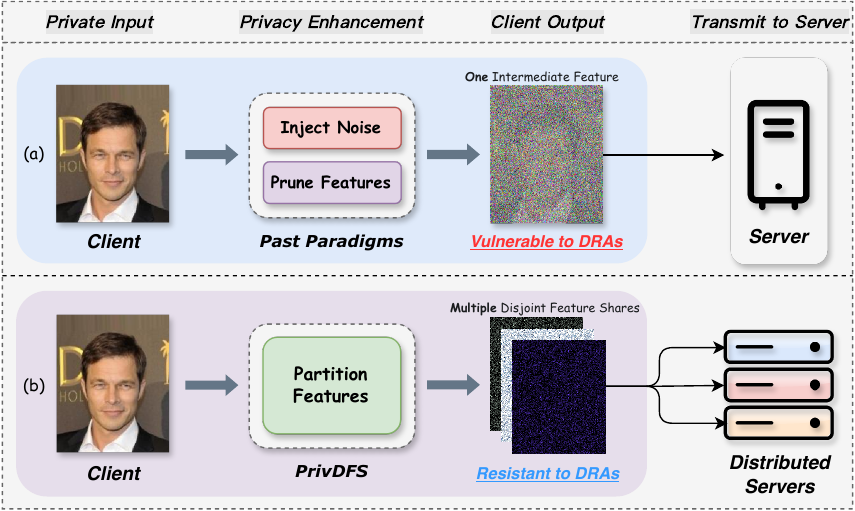}
    \caption{Comparison of architectural paradigms: (a) Conventional single-server split inference. (b) PrivDFS.}
    \label{fig:paradigm}
\end{figure}

Existing PI solutions fall into two broad categories: \emph{cryptographic methods} and \emph{split inference (SI)}. Cryptographic PI is typically built on Homomorphic Encryption (HE)~\cite{DBLP:conf/nips/XuWW024,DBLP:conf/iccv/PengH0LWWZX0GMW23,DBLP:journals/tc/ChengXLZGZS23} or Secure Multi-Party Computation (SMPC)~\cite{10824864,DBLP:conf/uss/DiaaFHDEK0LMOAK24}. While these methods provide strong, formally provable security guarantees, they incur substantial computational and communication overhead, making them difficult to deploy in latency-sensitive inference scenarios~\cite{DBLP:conf/asplos/GarimellaGJGR23}. In contrast, split inference~\cite{DBLP:conf/asplos/KangHGRMMT17,DBLP:journals/corr/abs-1812-00564} offers a more efficient and practical alternative by partitioning the model between the client and the cloud. The client executes the early layers of the model to obtain an intermediate feature representation (often called \emph{smashed data}), which is then sent to the server to complete the remaining computation. Unfortunately, SI inherently weakens privacy guarantees for client inputs: the intermediate features are transmitted in plaintext and can be directly observed by the cloud server, introducing a critical privacy vulnerability~\cite{liu2025data,DBLP:conf/ndss/0003LWJXO25,DBLP:conf/eccv/SinghSSMCYR24}. Data Reconstruction Attacks (DRAs) can exploit this exposed smashed data to reconstruct private inputs~\cite{DBLP:conf/ccs/FredriksonJR15}. Such attacks have rapidly evolved from early optimization-based methods with handcrafted priors~\cite{DBLP:conf/acsac/HeZL19,DBLP:conf/cvpr/UlyanovVL18,RUDIN1992259} to powerful generative inverters based on GANs~\cite{DBLP:conf/nips/LiYLWHYX23,DBLP:conf/eccv/QiuFYCQX24} and diffusion models~\cite{zhang2025unlocking,DBLP:conf/cvpr/ChenLZL0B22,pmlr-v267-lei25a}, which can produce highly accurate reconstructions from \textbf{a single holistic representation}. A natural countermeasure within the SI paradigm is to move the cut point deeper into the network so that the server receives more abstract features. However, this strategy substantially increases the computational burden on the client, undermining one of the main motivations for offloading inference to the cloud server.

To enhance privacy within the SI framework without relying solely on deeper splits, a line of work has proposed defenses that explicitly perturb or suppress the smashed data, as illustrated in Figure~\ref{fig:paradigm}. One family of methods injects noise into the intermediate features~\cite{DBLP:journals/tetci/PhamPC25,DBLP:conf/icdm/Vepakomma0GR20,DBLP:conf/asplos/MireshghallahTR20,DBLP:conf/cvpr/0005CGZV0R21,DBLP:conf/ndss/LiZ00Z24}, while another prunes feature channels or dimensions deemed less relevant to the downstream task~\cite{DBLP:conf/mm/BiHL0C024,DBLP:conf/www/MireshghallahTJ21}. In principle, both strategies aim to preserve information that is relevant to the primary task while discarding as much other information as possible. However, feature entanglement~\cite{DBLP:conf/mm/BiHL0C024} makes it extremely difficult to precisely isolate task-irrelevant components: the same feature dimensions often encode both high-level semantic information useful for prediction and low-level details exploitable by DRAs. As a result, stronger noise injection or more aggressive pruning can indeed improve privacy, but at the cost of a dramatic drop in model utility.

These limitations call for a PI framework that can jointly achieve utility, privacy, and efficiency, without relying on heavier client computation or feature suppression. To this end, we propose \textbf{PrivDFS}, a distributed feature-sharing framework for private inference inspired by \emph{secret sharing}~\cite{DBLP:journals/cacm/Shamir79}. Instead of directly perturbing or discarding parts of a single holistic representation, PrivDFS replaces it with \emph{feature shares}: a client-side Distributed Feature Sharing (DFS) module decomposes the smashed data into multiple, largely disjoint shares and distributes them to an honest-majority set of cloud servers. Each server only observes its own share and performs inference locally without inter-server communication, and a lightweight fusion module on the client aggregates the partial predictions to obtain the final output. This design fundamentally changes the attack surface for DRAs: even if a server is compromised (or a minority coalition), it only obtains a partial, fragmented view of the input, making accurate reconstruction difficult, while the complete set of shares collectively maintains a virtually intact representation for high-utility inference. Notably, PrivDFS preserves client efficiency: only a lightweight encoder, DFS, and fusion module run locally, making the framework practical for cloud-based image classification.

\noindent\textbf{Contributions.} Our main contributions are as follows:
\begin{itemize}
    \item We introduce \textbf{PrivDFS}, a new paradigm for private inference that replaces a single exposed representation with distributed feature shares across multiple honest-majority servers, substantially reducing the attack surface for DRAs.
    
    \item We develop a learnable \textbf{Distributed Feature Sharing (DFS)} mechanism that automatically partitions smashed data into sparse and largely non-overlapping shares, effectively limiting per-server visibility while preserving task-relevant information.
    
    \item We show that PrivDFS is architecture-general and instantiate \textbf{PrivDFS-ViT} to demonstrate that distributed feature sharing naturally extends from convolutional backbones to modern transformer-based models.
    \item Extensive experiments on CIFAR-10/100, CelebA, and ImageNet-1K show that PrivDFS achieves a state-of-the-art privacy-utility-efficiency trade-off with minimal accuracy degradation.
\end{itemize}

%% file: 2_background.tex
\section{Threat Model}
\label{sec:threat_model}

\begin{figure*}[h!]
    \centering
    \includegraphics[width=0.9\textwidth]{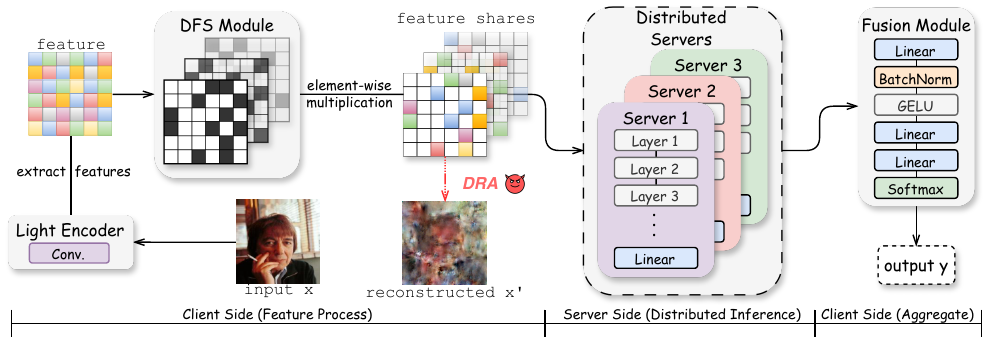}
    \caption{Illustration of the PrivDFS framework for an example configuration with $K=3$ servers.}    
    \label{fig:PrivDFS_framework}
\end{figure*}

\textbf{Scenario.}
We focus on protecting the client’s input privacy in cloud-based \emph{image classification}. The client holds a private input image \(x\) and computes an intermediate representation \(z\) through a lightweight local encoder \(M_c^{\mathrm{enc}}\). It then applies the DFS module to fragment \(z\) into multiple feature shares \(\{z_i\}_{i=1}^K\). Each share \(z_i\) is sent to a different cloud server, which only processes its own share and returns a partial prediction \(y_i\) to the client. We assume an honest-majority setting: at most a strict minority of the servers can be compromised. The adversary controls this coalition and observes all shares and intermediate activations held by the compromised servers (i.e., \(\{z_i\}_{i \in \mathcal{C}}\)), and aims to reconstruct the private input from this partial, distributed view.

\textbf{Adversary Knowledge and Capability.}
We assume a powerful gray-box adversary that controls a strict minority of servers indexed by \(\mathcal{C} \subset \{1,\dots,K\}\). The adversary knows the architectures and parameters of the server-side models \(\{M_{s,i}\}_{i \in \mathcal{C}}\), but has no access to the parameters of the client-side encoder \(M_c\) or the computations on honest servers. The coalition can aggregate all feature shares and activations it observes, i.e., \(\{z_i\}_{i \in \mathcal{C}}\), and aims to train an inversion model \(\mathcal{A}\) such that \(\mathcal{A}(\{z_i\}_{i \in \mathcal{C}}) \approx x\). To this end, it constructs a training set of input–representation pairs by querying the deployed system and recording \(\big(x, \{z_i\}_{i \in \mathcal{C}}\big)\) for inputs drawn from the data distribution~\cite{DBLP:conf/acsac/HeZL19}. For a rigorous and conservative evaluation, we assume that the adversary can use the same public dataset as the one used to train the victim model when training its inversion model.

%% file: 3_method.tex
\section{Proposed Framework: PrivDFS}
\subsection{Framework Overview}

As illustrated in Figure~\ref{fig:PrivDFS_framework}, inference under PrivDFS proceeds in three stages along the client--server pipeline. The client first applies the lightweight local encoder \(M_c^{\mathrm{enc}}\) to the raw input \(x\) and obtains a compact intermediate representation \(z = M_c^{\mathrm{enc}}(x)\). This smashed data is then passed to the Distributed Feature Sharing (DFS) module, which decomposes \(z\) into \(K\) feature shares \(\{z_i\}_{i=1}^K\). Concretely, DFS module uses learnable binary sparsity masks to select and route different subsets of features to the distributed servers, yielding shares that are approximately sparse, disjoint, and balanced in size. Each share \(z_i\) is transmitted to a distinct cloud server \(i\), which hosts a server-side sub-model \(M_{s,i}\) that performs partial inference and returns a partial prediction \(y_i = M_{s,i}(z_i)\) to the client. Finally, a lightweight client fusion module \(M_c^{\mathrm{fus}}\) aggregates the partial predictions into the final output \(y = M_c^{\mathrm{fus}}(\{y_i\}_{i=1}^K)\). Throughout this process, the raw input \(x\) and final prediction \(y\) never leave the client device: the servers only observe their own feature share and associated activations.

PrivDFS enforces privacy by \emph{distribution rather than perturbation}. Instead of injecting noise into a single holistic representation or aggressively pruning it, the DFS module structurally fragments the smashed data into multiple feature shares using its learned sparsity masks. This design is guided by three objectives. First, each individual share should be deliberately incomplete and semantically limited, so that any single server---or any strict minority coalition of servers---only gains a partial view of the underlying input. Second, when the partial predictions derived from all shares are aggregated by the fusion module, they should collectively preserve task-relevant information and support high-utility inference, keeping the accuracy close to that of non-private split inference. Third, the client-side computation, including the encoder, DFS, and fusion modules, must remain lightweight enough to be practical for resource-constrained devices.

By replacing a single exposed intermediate representation with a set of sparse, distributed feature shares, PrivDFS fundamentally reshapes the attack surface compared to conventional split inference. An adversary that compromises a strict minority of servers, as assumed in our threat model, can only access a subset of the feature shares and their corresponding activations, which substantially increases the difficulty of training effective data reconstruction models from this fragmented view. At the same time, the honest-majority collection of servers still operates on a near-complete representation in aggregate, enabling accurate predictions with minimal additional latency or client overhead. Our experimental results in Section~\ref{evaluation} demonstrate that this distributed feature-sharing design yields a favorable privacy--utility--efficiency trade-off under strong diffusion-based reconstruction attacks.

\subsection{Module Design: DFS and Fusion}

\textbf{DFS Module.}
Let \(z \in \mathbb{R}^{C \times H \times W}\) denote the intermediate feature map produced by the client encoder, where \(C\) is the number of channels and \(H \times W\) is the spatial resolution (for transformer-based backbones, we reshape token embeddings into an equivalent 2D tensor with the same notation). The DFS module associates each of the \(K\) server branches with a learnable binary mask \(\mathrm{M}_i \in \{0,1\}^{C \times H \times W}\) of the same shape as \(z\). Each mask selects a subset of feature locations for branch \(i\) via element-wise multiplication, \(z_i = z \odot \mathrm{M}_i\), where \(\odot\) denotes the Hadamard product. The resulting \(z_i\) is the feature share sent to server \(i\).

A technical challenge is that binary masks are non-differentiable, preventing direct optimization by backpropagation. We address this by combining the Gumbel--Softmax relaxation with a Straight-Through Estimator (STE)~\cite{DBLP:conf/iclr/JangGP17,DBLP:journals/corr/BengioLC13}. During training, each \(\mathrm{M}_i\) is parameterized by real-valued logits, which are passed through a Gumbel--Softmax sampler to obtain a continuous approximation in \([0,1]^{C \times H \times W}\). In the forward pass, we apply a hard threshold to obtain a binary mask used to compute \(z_i\). In the backward pass, the STE treats this thresholding operation as the identity, allowing gradients to propagate as if the continuous relaxation had been used. This standard trick enables end-to-end learning of discrete, hard feature-selection masks.

To guide the masks toward a structured feature partition that is suitable for distributed inference, we optimize them jointly with the server-side sub-models under the supervision of the primary task loss and several auxiliary regularizers. Intuitively, we would like (i) each branch to receive only a small subset of features, (ii) different branches to focus on largely non-overlapping subsets, and (iii) the union of all branches to still cover a sufficiently rich set of features so that the fused prediction remains accurate. Such a partitioning enables the model to preserve task-relevant information in aggregate while reducing the amount of information exposed to any individual branch (or strict minority coalition), which is beneficial from a privacy perspective.

Formally, we first introduce a \textbf{sparsity constraint} that controls the fraction of features selected by each mask. Let
\begin{equation}
    \bar{\mathrm{M}}_i = \frac{1}{C H W} \sum_{c,h,w} \mathrm{M}_i[c,h,w]
\end{equation}
denote the average activation of mask \(\mathrm{M}_i\). We set the target sparsity to \(\rho = 1/K\) in Eq.~\eqref{eq:sparsity_loss} so that each branch observes only a small, incomplete share of the overall feature map. To this end, we ensure \(\bar{\mathrm{M}}_i\) to match a target sparsity level \(\rho \in (0,1]\) via
\begin{equation}
    \mathcal{L}_{\text{sparsity}} = \sum_{i=1}^{K} \left( \bar{\mathrm{M}}_i - \rho \right)^2.
    \label{eq:sparsity_loss}
\end{equation}

Second, we ensure different branches to attend to distinct subsets of features through a \textbf{diversity constraint}. Specifically, we penalize the cosine similarity between any pair of masks:
\begin{equation}
\mathcal{L}_{\text{diversity}}
= \sum_{1 \le i < j \le K}
\frac{\langle \mathrm{M}_i, \mathrm{M}_j \rangle_F}
{\|\mathrm{M}_i\|_F \, \|\mathrm{M}_j\|_F + \epsilon},
\label{eq:diversity_loss}
\end{equation}
where \(\langle \cdot, \cdot \rangle_F\) is the Frobenius inner product (i.e., \(\langle A, B \rangle_F = \sum_{c,h,w} a_{c,h,w} b_{c,h,w}\)), \(\|\cdot\|_F\) is the Frobenius norm, and \(\epsilon\) is a small constant for numerical stability. Minimizing \(\mathcal{L}_{\text{diversity}}\) ensures each branch to specialize on a distinct feature subspace, reducing overlap between shares.

Third, we introduce a \textbf{balance constraint} that aligns the sparsity levels of different branches. Using the per-branch sparsities \(\{\bar{\mathrm{M}}_i\}_{i=1}^K\), we define
\begin{equation}
\mathcal{L}_{\text{balance}} = \frac{1}{K}\sum_{i=1}^{K} (\bar{\mathrm{M}}_i - \mu_{\bar{\mathrm{M}}})^2,
\label{eq:balance_loss}
\end{equation}
where \(\mu_{\bar{\mathrm{M}}} = \frac{1}{K} \sum_{i=1}^{K} \bar{\mathrm{M}}_i\) is the mean sparsity across branches. This term keeps the sparsity levels across branches closely aligned, so that the masks remain approximately balanced and no server systematically receives substantially more features than the others.

The overall training objective is a weighted sum of the primary task loss \(\mathcal{L}_{\text{task}}\) and the above structural regularizers:
\begin{equation}
\mathcal{L} = \mathcal{L}_{\text{task}} 
+ \lambda_s \mathcal{L}_{\text{sparsity}} 
+ \lambda_b \mathcal{L}_{\text{balance}} 
+ \lambda_d \mathcal{L}_{\text{diversity}},
\label{eq:total_loss}
\end{equation}
where \(\lambda_s\), \(\lambda_b\), and \(\lambda_d\) are hyperparameters controlling the strength of each regularizer. These terms shape the space of admissible feature partitions rather than explicitly trading off against accuracy: as \(\mathcal{L}_{\text{sparsity}}\), \(\mathcal{L}_{\text{balance}}\), and \(\mathcal{L}_{\text{diversity}}\) are driven toward zero, the model learns masks that are sparse, balanced, and diverse while still minimizing \(\mathcal{L}_{\text{task}}\). In practice, this produces a partition of the intermediate features that preserves high task performance after fusion, with near-disjoint masks that collectively cover almost all features while reducing the amount and redundancy of information exposed to any individual server.

\textbf{Fusion Module.}
Each server \(i\) receives its feature share \(z_i\) and produces a partial prediction \(y_i = M_{s,i}(z_i)\). Given the collection of partial predictions \(\{y_i\}_{i=1}^K\), the client-side fusion nodule \(M_c^{\mathrm{fus}}\) aggregates them into the final output
\begin{equation}
    y = M_c^{\mathrm{fus}}([y_1 \Vert y_2 \Vert \cdots \Vert y_K]).
\end{equation}

In our implementation, \(M_c^{\mathrm{fus}}\) is instantiated as a shallow multilayer perceptron followed by a softmax layer for classification. Leveraging the structured feature partitioning enforced by DFS, this low-capacity fusion module is sufficient to aggregate the signals from different branches while incurring minimal additional client-side overhead.

Together, the DFS and fusion modules are designed so that each feature share exposes only a sparse, partial view of the intermediate representation, while the aggregation of server-side partial predictions on the client yields an output that remains suitable for accurate inference. By structurally partitioning features across branches and using only a low-capacity fusion nodule on the client, PrivDFS reduces the amount of information accessible to any individual server or strict minority coalition, without introducing substantial additional client-side computation.

\subsection{Adaptation to Vision Transformers}
To demonstrate that PrivDFS is not limited to convolutional backbones, we instantiate it on the Vision Transformer (ViT) architecture~\cite{DBLP:conf/iclr/DosovitskiyB0WZ21}. 
Although both CNNs and ViTs process images, their intermediate representations differ fundamentally: CNNs use spatial feature maps \(z \in \mathbb{R}^{C \times H \times W}\), whereas ViTs operate on sequences of patch tokens \(z \in \mathbb{R}^{N \times D}\). 
This structural difference requires adapting DFS from a channel–spatial masking mechanism to a token–embedding masking mechanism. 
Importantly, DFS does not rely on convolution-specific structure; it only requires a structured intermediate tensor, whether a spatial grid or a token matrix, making the extension to ViTs conceptually natural.

A standard ViT divides an input image into non-overlapping patches, projects each patch into an embedding, and adds positional encodings to form a token sequence \(z \in \mathbb{R}^{N \times D}\), which is then processed by Transformer encoder blocks. 
Compared to the hierarchical spatial representation of CNNs, this sequence-based representation organizes information across tokens instead of spatial locations, meaning that DFS must mask along token and embedding dimensions rather than channels and spatial positions.

In PrivDFS-ViT, the client executes the patch embedding layer and the first \(L\) Transformer encoder blocks. These early blocks enrich raw patch embeddings with local and short-range dependencies, producing an intermediate sequence that remains lightweight to compute while retaining sufficient semantic structure for downstream inference. 
The DFS module then generates \(K\) feature shares by applying a learnable binary mask \(\mathrm{M}_i \in \{0,1\}^{N \times D}\) to the intermediate sequence: $z_i = z \odot \mathrm{M}_i,$ where \(\odot\) denotes Hadamard product. Each masked sequence \(z_i\) is sent to a dedicated server-side ViT sub-model \(M_{s,i}\), and the resulting partial predictions are fused on the client as described previously.

The masks \(\mathrm{M}_i\) are trained exactly as in the convolutional case: using a Gumbel--Softmax relaxation with a Straight-Through Estimator to enable end-to-end optimization, and reusing the sparsity, diversity, and balance regularizers (Eqs.~\eqref{eq:sparsity_loss}--\eqref{eq:balance_loss}) with the mask shape changed from \(C \times H \times W\) to \(N \times D\). 
Thus, the DFS design and training objective remain unchanged, demonstrating that PrivDFS is architecture-agnostic.

\textbf{Effect on Attention and Utility.}
Because DFS is applied after the first \(L\) encoder blocks, the tokens entering the masking stage have already been context-enriched by self-attention. 
During training, each server-side sub-model is jointly optimized with its corresponding mask, allowing the attention layers to adapt to sparsified token sequences rather than being unexpectedly disrupted at inference time. 
From an optimization viewpoint, DFS behaves like a learned and structured form of representation-level dropout: different branches receive complementary subsets of tokens and embedding dimensions, while the sparsity, balance, and diversity regularizers ensure that their union preserves most task-relevant information. 
Consequently, each branch alone provides only a partial and semantically incomplete view of the input—limiting the reconstructive power of any single server—while the fused predictions on the client retain strong accuracy. 
As shown in Section~\ref{evaluation}, this adaptation enables PrivDFS-ViT to achieve improved privacy--utility--efficiency trade-offs under strong DRAs.

%% file: 4_evaluation.tex
\section{Evaluation}
\label{evaluation}

\begin{table*}[htbp]
\centering
\small
\setlength{\tabcolsep}{3pt} 
\begin{tabular}{@{}c|cccc|cccc|cccc@{}}
\toprule
\multirow{2}{*}{\textbf{Method}} & \multicolumn{4}{c|}{\textbf{CIFAR-10}} & \multicolumn{4}{c|}{\textbf{CIFAR-100}} & \multicolumn{4}{c}{\textbf{CelebA}} \\
\cmidrule(lr){2-5} \cmidrule(lr){6-9} \cmidrule(lr){10-13}
& \textbf{Acc} $\uparrow$ & \textbf{PSNR} $\downarrow$ & \textbf{SSIM} $\downarrow$ & \textbf{LPIPS} $\uparrow$ 
& \textbf{Acc} $\uparrow$ & \textbf{PSNR} $\downarrow$ & \textbf{SSIM} $\downarrow$ & \textbf{LPIPS} $\uparrow$
& \textbf{Acc} $\uparrow$ & \textbf{PSNR} $\downarrow$ & \textbf{SSIM} $\downarrow$ & \textbf{LPIPS} $\uparrow$ \\
\midrule
SI~\cite{DBLP:conf/asplos/KangHGRMMT17} & \underline{95.79\%} & 23.25 & 0.963 & 0.008
& \underline{79.36\%} & 21.36 & 0.923 & 0.012
& \underline{91.31\%} & 20.69 & 0.838 & 0.049 \\
DP~\cite{DBLP:journals/tetci/PhamPC25} & 86.23\% & 18.23 & 0.791 & 0.025
& 68.27\% & 17.65 & 0.772 & 0.033
& 89.26\% & 16.27 & 0.638 & 0.224 \\
PriFU~\cite{DBLP:conf/mm/BiHL0C024} & 82.59\% & 17.76 & 0.763 & 0.033
& 52.93\% & 17.24 & 0.765 & 0.034
& 90.29\% & 16.18 & 0.618 & 0.225 \\
VIM~\cite{DBLP:conf/ndss/LiZ00Z24} & 89.92\% & 16.51 & 0.712 & 0.050
& 70.38\% & 16.22 & 0.698 & 0.061
& 89.87\% & 14.89 & 0.598 & 0.245 \\
DISCO~\cite{DBLP:conf/cvpr/0005CGZV0R21} & 83.72\% & 19.78 & 0.815 & 0.018
& 63.85\% & 18.33 & 0.796 & 0.027
& 88.54\% & 17.59 & 0.702 & 0.199 \\
Cloak~\cite{DBLP:conf/www/MireshghallahTJ21} & 92.91\% & 18.31 & 0.802 & 0.019
& 71.80\% & 17.88 & 0.781 & 0.030
& 90.20\% & 16.58 & 0.654 & 0.218 \\
NoPeek~\cite{DBLP:conf/icdm/Vepakomma0GR20} & 80.23\% & 18.65 & 0.809 & 0.018
& 59.91\% & 17.95 & 0.786 & 0.029
& 89.35\% & 17.21 & 0.687 & 0.205 \\
Shredder~\cite{DBLP:conf/asplos/MireshghallahTR20} & 92.35\% & 17.59 & 0.782 & 0.035
& 73.18\% & 17.24 & 0.765 & 0.041
& 90.76\% & 15.46 & 0.615 & 0.230 \\
NCS & 95.21\% & 22.04 & 0.888 & 0.015
& 78.85\% & 20.21 & 0.853 & 0.019
& 91.27\% & 18.99 & 0.710 & 0.194 \\
\midrule
\textbf{PrivDFS(33\%)} &  \textbf{94.96\%} & \textbf{12.72} & \textbf{0.260} & \textbf{0.164}
& \textbf{78.15\%} & \textbf{12.83} & \textbf{0.281} & \textbf{0.162}
& \textbf{90.83\%} & \textbf{9.82} & \textbf{0.128} & \textbf{0.556} \\
\textbf{PrivDFS(50\%)} 
& \textbf{94.93\%} & \textbf{13.72} & \textbf{0.262} & \textbf{0.117}
& \textbf{77.98\%} & \textbf{13.92} & \textbf{0.224} & \textbf{0.154}
& \textbf{90.71\%} & \textbf{9.93} & \textbf{0.133} & \textbf{0.553} \\
\bottomrule
\end{tabular}
\caption{Accuracy and privacy metrics under DRAs on CIFAR‑10, CIFAR-100 and CelebA. For PrivDFS, 33\% and 50\% correspond to the fraction of compromised servers (i.e., 1 of 3 servers and 3 of 6 servers, respectively).}
\label{tab:privacy_comparison_extended}
\end{table*}

We evaluate our framework along three aspects:
(1) \emph{utility}, measured by classification accuracy on multiple benchmark datasets;
(2) \emph{privacy}, assessed using state-of-the-art Data Reconstruction Attacks to quantify the fidelity of recovered inputs; and
(3) \emph{efficiency}, evaluated in terms of client-side computation and end-to-end latency.
All models are implemented in PyTorch~\cite{DBLP:conf/nips/PaszkeGMLBCKLGA19}. Unless otherwise specified, the number of server branches is set to $K = 3$. All experiments are repeated 10 times, and we report the mean of the results. Experiments are conducted on a high-performance cluster equipped with Intel Xeon Gold 6330 CPUs, 1\,TB RAM, and 8 NVIDIA RTX 4090 GPUs.

\subsection{Experimental Settings}

\textbf{Datasets and Models.}
We evaluate our framework on three standard benchmarks---CIFAR-10, CIFAR-100~\cite{krizhevsky2009learning}, and CelebA~\cite{DBLP:conf/iccv/LiuLWT15}---for multi-class and multi-attribute classification, and further extend to the large-scale ImageNet-1K dataset~\cite{imagenet15russakovsky} to assess scalability.
The client-side encoder $M_c^{\mathrm{enc}}$ is deliberately lightweight, consisting of a single convolutional layer that extracts low-level features.
For our primary experiments on CIFAR-10, CIFAR-100, and CelebA, each server branch $M_{S,i}$ is constructed from the main body of a ResNet-18~\cite{DBLP:conf/cvpr/HeZRS16} backbone, comprising its eight standard BasicBlocks.
To demonstrate performance on more challenging tasks, we scale up our approach for ImageNet-1K, where each server branch is composed of the sixteen Bottleneck blocks from a ResNet-50 backbone. 
Furthermore, to validate the general applicability of PrivDFS beyond convolutional architectures, we extend our evaluation to ViT.
Specifically, we assess a ViT-Small~\cite{DBLP:conf/iclr/DosovitskiyB0WZ21} model on CIFAR-10.
In this configuration, the client-side model consists of initial patch embedding layers and the first three Transformer encoder blocks (out of a total of twelve), with the remaining layers distributed among the server branches.

\textbf{Adversary Setting.}
To rigorously evaluate the privacy guarantees of PrivDFS, we adopt a powerful, state-of-the-art DRA as our primary adversary.
Diffusion models have recently emerged as the gold standard for reconstruction attacks, consistently outperforming GAN- and VAE-based approaches in their ability to recover fine-grained visual details from highly compressed or partial representations~\cite{zhang2025unlocking,pmlr-v267-lei25a,DBLP:conf/cvpr/ChenLZL0B22}.
Their iterative denoising process, guided by strong learned priors, makes them exceptionally effective at exploiting even weak or fragmented feature cues, such as those present in our distributed shares, and thus they constitute a natural worst-case attacker. Specifically, we employ the diffusion-based DIA~\cite{DBLP:conf/cvpr/ChenLZL0B22} as our reference attack.
This choice aligns with our threat model, in which an adversary trains a dedicated inversion model to map the received intermediate features back to the private inputs.
By evaluating against DIA, we test PrivDFS under a highly capable adversary that represents the current frontier in DRAs.

\textbf{Baselines.}
We compare our framework against a comprehensive set of baselines spanning two major defense strategies, plus two fundamental controls:
(i) \emph{perturbation-based defenses}, which protect privacy by adding noise or otherwise obfuscating the data, including methods that operate on intermediate features such as \textbf{Differential Privacy (DP)}~\cite{DBLP:journals/tetci/PhamPC25}, information-theoretic noise addition like \textbf{NoPeek}~\cite{DBLP:conf/icdm/Vepakomma0GR20}, learned noise injection as in \textbf{Shredder}~\cite{DBLP:conf/asplos/MireshghallahTR20}, and adversarially trained channel permutation as in \textbf{DISCO}~\cite{DBLP:conf/cvpr/0005CGZV0R21}, as well as the input-level defense \textbf{VIM}~\cite{DBLP:conf/ndss/LiZ00Z24}, which obfuscates the raw image by shuffling pixels in spatial and chromatic domains;
(ii) \emph{feature-pruning defenses}, which aim to reduce information leakage by removing or masking task-irrelevant components of the intermediate representation, including \textbf{PriFU}~\cite{DBLP:conf/mm/BiHL0C024}, which disentangles and discards private attributes, and \textbf{Cloak}~\cite{DBLP:conf/www/MireshghallahTJ21}, which preserves only a minimal subset of essential features;
and (iii) \emph{control baselines}, where we use standard \textbf{Split Inference}~\cite{DBLP:conf/asplos/KangHGRMMT17} as a utility upper bound (with no defense) and a \textbf{Naive Channel Split (NCS)} ablation to show that the learned feature distribution provided by our DFS module is superior to a simple, non-learned partitioning of channels.

\textbf{Evaluation Metrics.}
We evaluate each method on three dimensions:
(1) \emph{Utility}, as Top-1 classification accuracy (\%$\uparrow$);
(2) \emph{Efficiency}, as client-side FLOPs ($\downarrow$) and end-to-end latency ($ms$ $\downarrow$); and
(3) \emph{Privacy}, via the reconstruction fidelity of a diffusion-based attacker, measured by
PSNR ($\downarrow$)~\cite{DBLP:conf/icpr/HoreZ10}, SSIM ($\downarrow$)~\cite{DBLP:journals/tip/WangBSS04}, and LPIPS ($\uparrow$)~\cite{DBLP:conf/cvpr/ZhangIESW18}.
Lower PSNR/SSIM and higher LPIPS indicate stronger privacy.

Full implementation details and hyperparameter settings are provided in Appendix~B.

\subsection{Experiments on Privacy}

\begin{figure*}[t]
    \centering
    \includegraphics[width=0.9\textwidth]{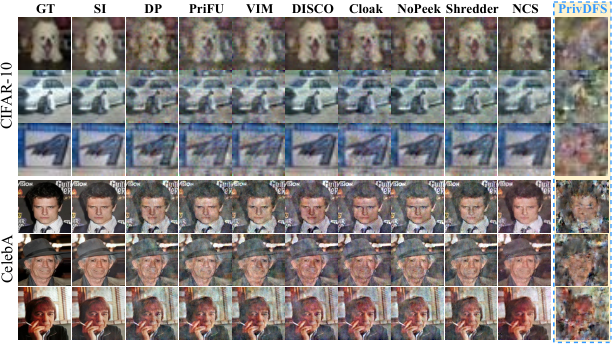}
    \caption{Qualitative reconstruction results under DRA on CIFAR-10 and CelebA. Columns correspond to different defense methods.}
    \label{fig:vis_DRAs}
\end{figure*}

Our primary privacy results are summarized in Table~\ref{tab:privacy_comparison_extended}.
The undefended SI baseline is highly vulnerable to DRAs, yielding high-fidelity reconstructions with high PSNR/SSIM and very low LPIPS (e.g., 23.25/0.963/0.008 on CIFAR-10).
Existing defenses reduce reconstruction quality to some extent but exhibit a pronounced privacy--utility trade-off: methods such as DP, PriFU, and NoPeek noticeably lower PSNR and SSIM yet incur large accuracy drops (e.g., PriFU at 52.93\% and NoPeek at 59.91\% on CIFAR-100).

In contrast, PrivDFS largely breaks this trade-off.
With three servers and one compromised server (33\%), PrivDFS achieves the strongest privacy across all datasets while keeping accuracy within 1--2 percentage points of SI.
On CIFAR-10, PrivDFS(33\%) reduces PSNR/SSIM from 23.25/0.963 (SI) to 12.72/0.260 and increases LPIPS from 0.008 to 0.164 at 94.96\% accuracy (vs.\ 95.79\% for SI), and similar improvements are observed on CIFAR-100 and CelebA.
Figure~\ref{fig:vis_DRAs} qualitatively confirms these trends: while baseline defenses often yield distorted but still recognizable reconstructions, those under PrivDFS become visually uninformative, indicating that the observed shares do not retain a coherent view of the input.
In Appendix~A, we further visualize how a single intermediate feature map is processed by DFS into multiple sparse feature shares across branches.
Even when stress-tested beyond the assumed threat model, PrivDFS remains effective even when up to 50\% of servers are compromised, with only a modest weakening of privacy metrics and negligible further impact on accuracy.
The NCS ablation further underscores the role of the DFS module.
Although NCS maintains accuracy close to SI, its privacy metrics lie much closer to the undefended baseline than to PrivDFS (e.g., PSNR 22.04 vs.\ 12.72 on CIFAR-10), showing that a simple, non-learned channel partition is insufficient.
The learned, structured fragmentation enforced by DFS is essential for obtaining strong privacy gains without sacrificing utility.

\paragraph{Scalability to ImageNet-1K.}
To assess scalability, we further evaluate PrivDFS on the more demanding ImageNet-1K benchmark with a ResNet-50 backbone.
On this 1000-way classification task, PrivDFS achieves a Top-1 accuracy of \textbf{71.13\%}, reasonably close to the \textbf{76.65\%} of the non-private SI baseline, indicating that our method retains high utility at ImageNet scale.
Under the diffusion-based DIA attack, PrivDFS yields reconstruction metrics of PSNR = 12.98, SSIM = 0.261, and LPIPS = 0.163, which fall in the same low-fidelity regime as our CIFAR-10/100 results in Table~\ref{tab:privacy_comparison_extended}.
Moreover, our results on CelebA, CIFAR-10, and CIFAR-100 already indicate that competing defenses exhibit rapidly increasing accuracy degradation as the label space grows, making them unlikely to remain competitive on this substantially more challenging benchmark.
Taken together, these findings suggest that PrivDFS is well suited for large-scale, real-world deployments that require both strong privacy and high utility.

\subsection{Experiments on Efficiency}

\textbf{Client-Side Cost.}
Table~\ref{tab:si_privdfs_efficiency} compares PrivDFS with SI configured to achieve a similar level of privacy by moving the cut after Block5 of ResNet-18.
This deep split substantially increases client computation: on CIFAR-10, SI requires 206.7M FLOPs, whereas PrivDFS attains comparable privacy (PSNR/SSIM/LPIPS of 12.72/0.260/0.164 vs.\ 12.96/0.279/0.158) with only 1.14M FLOPs, i.e., over 180$\times$ less.
On the higher-resolution CelebA dataset, the gap is nearly 250$\times$ (4.39G vs.\ 18.3M FLOPs) under similarly strong privacy.
These results show that PrivDFS can provide strong protection without pushing additional computation onto the client, effectively decoupling privacy from client cost in contrast to traditional split inference.

\begin{table}[htbp]
\centering
\small
\setlength{\tabcolsep}{4pt}  
\begin{tabular}{@{}cccccc@{}}
\toprule
\textbf{Dataset} & \textbf{Method} & \textbf{FLOPs} $\downarrow$ & \textbf{PSNR} $\downarrow$ & \textbf{SSIM} $\downarrow$ & \textbf{LPIPS} $\uparrow$ \\
\midrule
\multirow{2}{*}{CIFAR-10}
  & SI          & 206.730M & 12.96 & 0.279 & 0.158 \\
  & \textbf{PrivDFS} & \textbf{1.143M}          & \textbf{12.72} & \textbf{0.260} & \textbf{0.164} \\
\midrule
\multirow{2}{*}{CelebA}
  & SI          & 4.387G & 9.95 & 0.144 & 0.540 \\
  & \textbf{PrivDFS} & \textbf{18.285M}          & \textbf{9.82} & \textbf{0.128} & \textbf{0.556} \\
\bottomrule
\end{tabular}
\caption{Privacy vs. efficiency between SI and PrivDFS under matched privacy leakage levels.}
\label{tab:si_privdfs_efficiency}
\end{table}

\begin{table}[htbp]
\centering
\small
\setlength{\tabcolsep}{4pt} 
\renewcommand{\arraystretch}{1.1} 
\begin{tabularx}{\columnwidth}{@{}>{\centering\arraybackslash}m{2.6cm}|*{3}{>{\centering\arraybackslash}X}@{}}
\toprule
\textbf{Method} & \textbf{CIFAR-10} & \textbf{CIFAR-100} & \textbf{CelebA} \\
\midrule
SI~\cite{DBLP:conf/asplos/KangHGRMMT17} & 2.033 & 2.086 & 2.204 \\
DP~\cite{DBLP:journals/tetci/PhamPC25} & 2.742 & 2.812 & 2.951 \\
PriFU~\cite{DBLP:conf/mm/BiHL0C024} & 6.881 & 6.962 & 7.415 \\
DISCO~\cite{DBLP:conf/cvpr/0005CGZV0R21} & 8.325 & 8.510 & 8.962 \\
Cloak~\cite{DBLP:conf/www/MireshghallahTJ21} & 5.321 & 5.446 & 5.781 \\
NoPeek~\cite{DBLP:conf/icdm/Vepakomma0GR20} & 3.268 & 3.315 & 3.451 \\
Shredder~\cite{DBLP:conf/asplos/MireshghallahTR20} & 7.104 & 7.243 & 7.625 \\
FHE-based~\cite{DBLP:conf/uss/NamJLHOMP25} & 1826 & 1865 & 1921 \\
PrivDFS & 2.432 & 2.237 & 2.391 \\
\bottomrule
\end{tabularx}
\caption{End-to-end latency (ms) on CIFAR-10, CIFAR-100, and CelebA, measured on NVIDIA RTX 4090 GPUs.}
\label{tab:latency}
\end{table}

\textbf{End-to-end Latency.}
To assess real-world performance, we report total inference latency in Table~\ref{tab:latency}.
Here PrivDFS runs its three branches in parallel on three GPUs, while all other schemes use a single GPU.
Under this setting, PrivDFS incurs only a small overhead over the insecure SI baseline, remains comparable to other non-cryptographic defenses, and is still orders of magnitude faster than the FHE-based method.
These results indicate that PrivDFS offers a practical balance between privacy and responsiveness, remaining suitable for latency-sensitive applications.

\subsection{PrivDFS-ViT Performance}

\begin{figure}[htbp]
    \centering
    \includegraphics[width=\columnwidth]{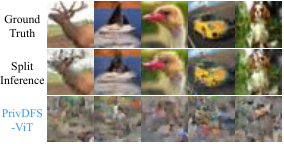} 
    \caption{Qualitative reconstruction results on CIFAR-10 under DRA, comparing standard SI with PrivDFS-ViT.}
    \label{fig:vit_vis}
\end{figure}

We adapt and evaluate PrivDFS-ViT (ViT-Small) architecture trained from scratch on CIFAR-10, without any external pretraining.
The results, presented in \textbf{Table~\ref{tab:vit}} and \textbf{Figure~\ref{fig:vit_vis}}, closely mirror our findings on CNN models.
Standard SI on ViT remains highly vulnerable, permitting clear reconstructions (PSNR of 20.287).
In contrast, PrivDFS-ViT effectively secures the ViT architecture, reducing the PSNR to 12.125 and producing visually unrecognizable reconstructions, while incurring only a negligible accuracy drop of 0.13\%.
These results confirm that the core principle of distributed feature sharing is model-agnostic and readily applicable to modern Transformer-based architectures.

\begin{table}[h]
    \centering
    \small
    \setlength{\tabcolsep}{6pt}
    \begin{tabular}{@{}c|cccc@{}}
        \toprule
        \textbf{Method} & \textbf{Acc} $\uparrow$ & \textbf{PSNR} $\downarrow$ & \textbf{SSIM} $\downarrow$ & \textbf{LPIPS} $\uparrow$\\
        \midrule
        SI & 88.46\% & 20.287 & 0.647 & 0.028 \\
        PrivDFS-ViT & 88.33\% & \textbf{12.125} & \textbf{0.095} & \textbf{0.172} \\
        \bottomrule
    \end{tabular}
    \caption{Quantitative comparison on ViT-Small (CIFAR-10, trained from scratch) under DRA.}    \label{tab:vit}
\end{table}

\subsection{Extended Experiments}

\begin{figure}[htbp]
    \centering
    \includegraphics[width=\columnwidth]{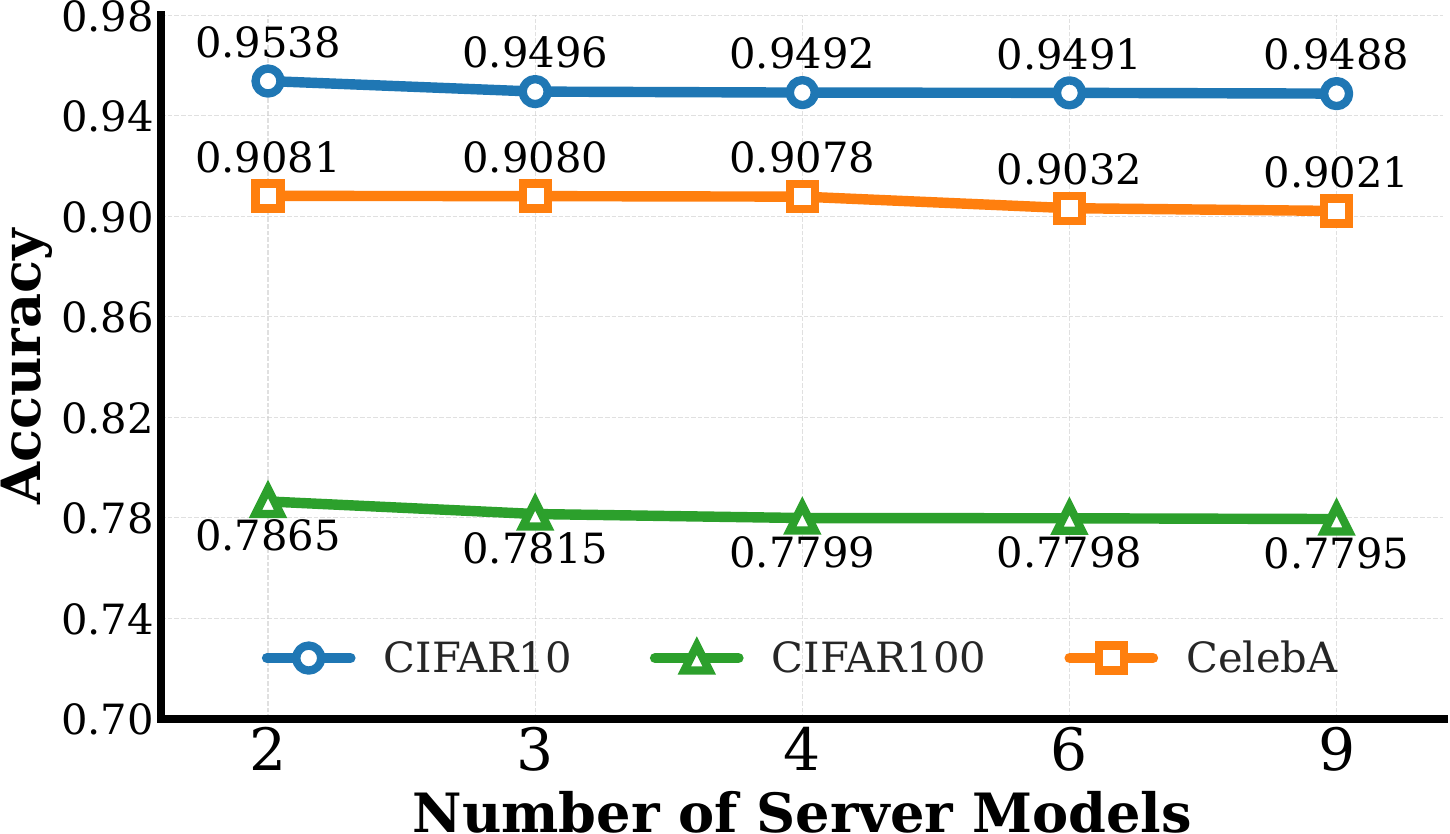} 
    \caption{Impact of the number of server models (\(K\)) on task accuracy: PrivDFS scales well with more branches.}
    \label{fig:ablation_n_servers}
\end{figure}

\paragraph{Scalability with Server Count.}
We evaluate the scalability of PrivDFS by varying the number of server branches (\(K\)) while keeping the client-side feature dimensionality fixed.
As shown in Figure~\ref{fig:ablation_n_servers}, accuracy decreases only very slightly as \(K\) increases and remains almost unchanged over a wide range of server counts, indicating that PrivDFS scales gracefully to different deployment configurations.

\paragraph{Inference Incapability of a Single Share.}
Finally, we verify the core security claim that no single branch is sufficiently informative
to support accurate prediction on its own.
When we directly classify using only one branch of PrivDFS on CIFAR-10,
the accuracies of the three branches are $10.7\%$, $9.6\%$, and $10.5\%$,
which are essentially indistinguishable from random guessing ($10\%$),
whereas fusing all three branches recovers a full accuracy of $94.9\%$.
This confirms that each individual feature share is semantically incomplete and does not, by itself,
provide a reliable predictive signal, thereby limiting both direct label inference
and reconstruction attacks that exploit partial predictive information like text-to-image diffusion models.

%% file: 5_conclusion.tex
\section{Conclusion and Future Work}  

We presented PrivDFS, a distributed feature-sharing framework that protects input privacy by fragmenting intermediate features across an honest-majority set of servers, so that any strict minority of compromised servers only observes incomplete feature shares. PrivDFS attains a favorable privacy--utility--efficiency trade-off: it offers strong resistance to diffusion-based DRAs with at most minor accuracy loss, while substantially reducing client-side computation compared to deeper split baselines. Experiments on benchmarks with both CNN and ViT architectures validate its effectiveness and architectural generality.

Looking ahead, several directions merit further exploration. While PrivDFS is client-efficient, the multi-server design introduces additional server-side and communication cost, motivating further optimization of both communication patterns and server-side computation. In addition, adapting the DFS module to pre-trained models could reduce the need for full end-to-end retraining when integrating PrivDFS into existing systems. Finally, extending distributed feature sharing beyond classification to tasks such as generative modeling could help make private inference more broadly applicable.